\documentclass[letterpaper, 10 pt, conference]{ieeeconf}  % Comment this line out if you need a4paper

\IEEEoverridecommandlockouts                              % This command is only needed if 
\usepackage{graphics} % for pdf, bitmapped graphics files
\usepackage{epsfig} % for postscript graphics files
\usepackage{mathptmx} % assumes new font selection scheme installed
\usepackage{times} % assumes new font selection scheme installed
\usepackage{booktabs}
\usepackage{amsmath} % assumes amsmath package installed
\usepackage{amssymb}  % assumes amsmath package installed
\usepackage[table, dvipsnames, svgnames, x11names]{xcolor} 
\usepackage{colortbl}
\usepackage{multirow}
\usepackage{marvosym}
\definecolor{mygray}{gray}{.9}
\usepackage{pifont}
\usepackage{cite}
\usepackage{CJKutf8}
\usepackage[utf8]{inputenc} % allow utf-8 input

\usepackage{bm}
\usepackage{color}
\usepackage{amsfonts}
\usepackage{algorithm}
\usepackage{algorithmic}
\usepackage{subfig}

\usepackage[colorlinks,
linkcolor=blue,
anchorcolor=blue,
citecolor=green,
urlcolor=Aqua,
backref=page]{hyperref}

\title{\LARGE \bf
Efficient Robotic Manipulation Through Offline-to-Online Reinforcement Learning and Goal-Aware State Information
}

\author{Jin Li$^{1}$, Xianyuan Zhan$^{1}$, Zixu Xiao$^{2}$, Guyue Zhou$^{1}$%
\thanks{$^{1}$Jin Li,  \textsuperscript{\Letter}Xianyuan Zhan, Guyue Zhou are with the Institute for AI Industry Research (AIR), Tsinghua University, 100084 Beijing, China
        {\tt\small \{lijin, zhanxianyuan, zhouguyue\}@air.tsinghua.edu.cn}}%
\thanks{$^{2}$Zixu Xiao is with the Department of Fluid Machinery and Engineering, Xi'an Jiaotong University, 710049 Xi'an, China
        {\tt\small smexiao@stu.xjtu.edu.cn}}%
}

\begin{document}

\maketitle
\thispagestyle{empty}

\pagestyle{empty}

%%%%%%%%%%%%%%%%%%%%%%%%%%%%%%%%%%%%%%%%%%%%%%%%%%%%%%%%%%%%%%%%%%%%%%%%%%%%%%%%
\begin{abstract}
% In this paper, we develop a new deep reinforcement learning (RL) framework 
% for efficient robotic manipulation.
% that greatly accelerate convergence speed of deep reinforcement learning in robotic manipulation tasks. Typical robotic manipulation system based on end-to-end deep reinforcement learning methods solves offline demonstration and online exploration as two separated process with no interleaving, and most of these systems are lack of a unified principle in the selection of visual observation states. Our work revolves around these two issues. First, we propose an offline-to-online transition method for deep reinforcement learning , which provides higher success rate during primary training stage. Additionally, we introduce a new method for observation selection in end-to-end visual complex tasks such as 'PickAndPlace' , which largely improve convergence speed of training process. Finally, we compare performance of our methods to current state-of-the-art methods in OpenAI-GYM-Fetch environment.

End-to-end learning robotic manipulation with high data efficiency is one of the key challenges in robotics. The latest methods that utilize human demonstration data and unsupervised representation learning has proven to be a promising direction to improve RL learning efficiency. The use of demonstration data also allows ``warming-up'' the RL policies using offline data with imitation learning or the recently emerged offline reinforcement learning algorithms. However, existing works often treat offline policy learning and online exploration as two separate processes, which are often accompanied by severe performance drop during the offline-to-online transition. Furthermore, many robotic manipulation tasks involve complex sub-task structures, which are very challenging to be solved in RL with sparse reward. In this work, we propose a unified offline-to-online RL framework that resolves the transition performance drop issue. Additionally, we introduce goal-aware state information to the RL agent, which can greatly reduce task complexity and accelerate policy learning. Combined with an advanced unsupervised representation learning module, our framework achieves great training efficiency and performance compared with the state-of-the-art methods in multiple robotic manipulation tasks.

\end{abstract}

%%%%%%%%%%%%%%%%%%%%%%%%%%%%%%%%%%%%%%%%%%%%%%%%%%%%%%%%%%%%%%%%%%%%%%%%%%%%%%%%
\section{INTRODUCTION}
In recent few years, deep reinforcement learning (RL) has seen great success in solving complex tasks such as games \cite{mnih2013playing,silver2017mastering,silver2018general} and robotic control \cite{schulman2015trust,levine2016end,schulman2017proximal,kalashnikov2018qt}.
The nice feature that RL can perform end-to-end learning with high-dimensional imagery input has made it a popular direction for dexterous robotic manipulation policy learning \cite{pinto2016supersizing, mahler2017dex,levine2018learning,gupta2018robot,andrychowicz2020learning,zhan2020framework}. 
Due to relatively high data collection cost on a real robot, developing high data-efficiency RL algorithms has become a key research focus in this area \cite{pinto2016supersizing, andrychowicz2020learning,zhan2020framework}.

Past studies on high data-efficiency RL-based robotic manipulation methods mainly follow three direction. The first stream of works focus on replacing the sparse reward in robotic manipulation tasks with dense reward \cite{wu2020learning} or introduce special reward structures \cite{cabi2019scaling,camacho2020disentangled}. Using more informative task reward can effectively reduce task complexity and lead to faster learning. However, this also has the drawback of involving heavy human engineering in the reward design, which loses generalizability across tasks as compared with the simple sparse reward. Another line of research is through Sim2Real, which learns an RL agents in simulation and then transfer to real world \cite{pinto2017asymmetric, andrychowicz2020learning, peng2018sim}. The downsides of this approach are high-variance in the learned polices and extensive computation involved due to training with domain randomization \cite{tobin2017domain}.

A more recent and effective direction to address the data-efficiency issue is through collecting small-scale real-world expert demonstration data for representation learning as well as policy pre-training via imitation learning \cite{schaal1999imitation, hussein2017imitation}, and then training the RL agent either in simulation or real world. This approach has achieved state-of-the-art performance and data-efficiency in various robotic manipulation tasks \cite{zhan2020framework}. The state representation is learned from real-world data, which commits lower error during Sim2Real adaptation. The pre-trained policy also facilitate faster online RL training.

\begin{figure}[t]
		\centering
		\subfloat[Impact of representation learning with expert and non-expert (20\% success rate) demonstrations]{
			\includegraphics[width=0.45\columnwidth]{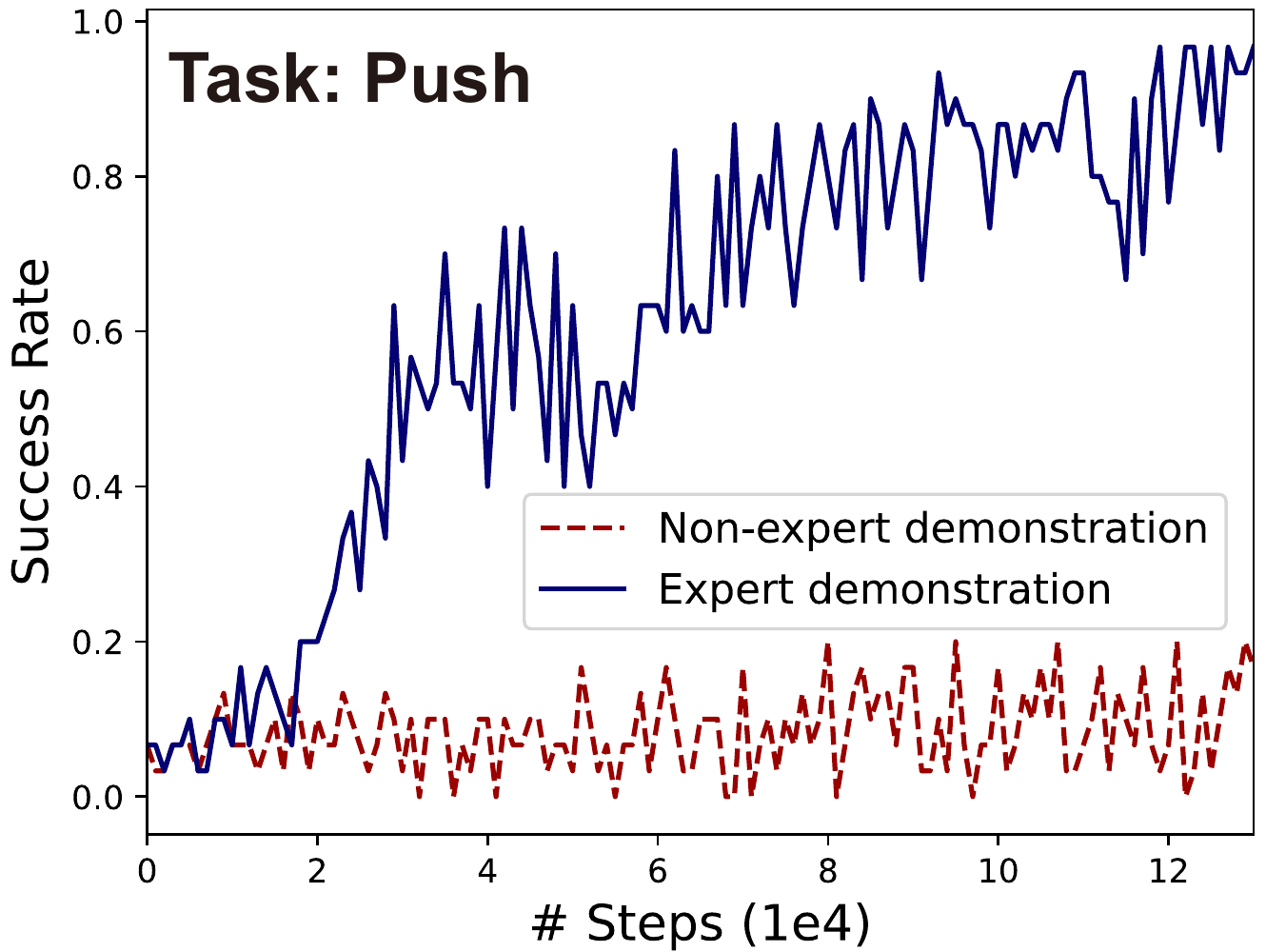}
			\label{fig:intro-1}
		}\,
		\subfloat[Impact of additional goal-stage aware observations and non-goal related observations]{
			\includegraphics[width=0.45\columnwidth]{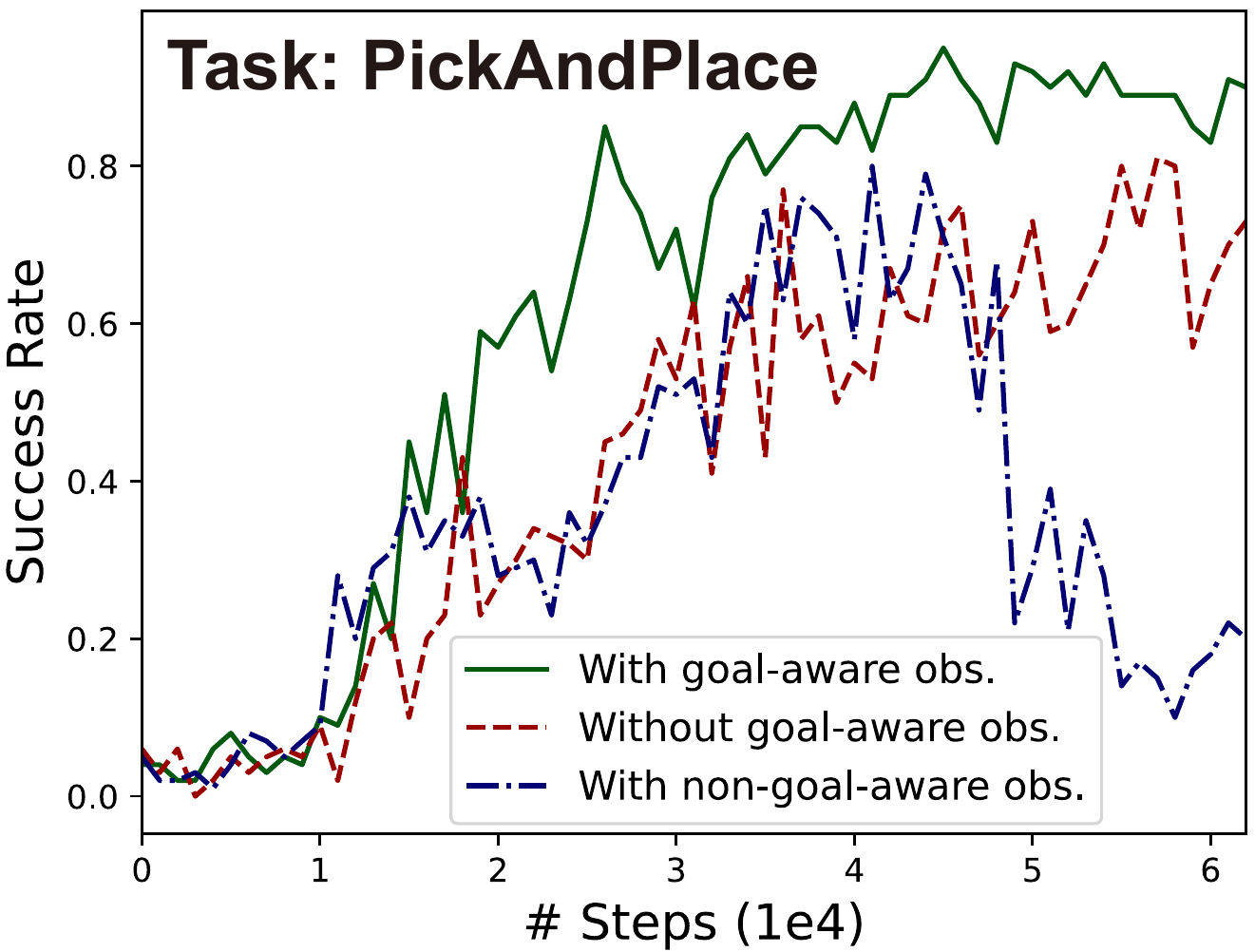}
			\label{fig:intro-2}
		} \\
		\subfloat[Offline-to-online performance gap]{
			\includegraphics[width=0.48\columnwidth]{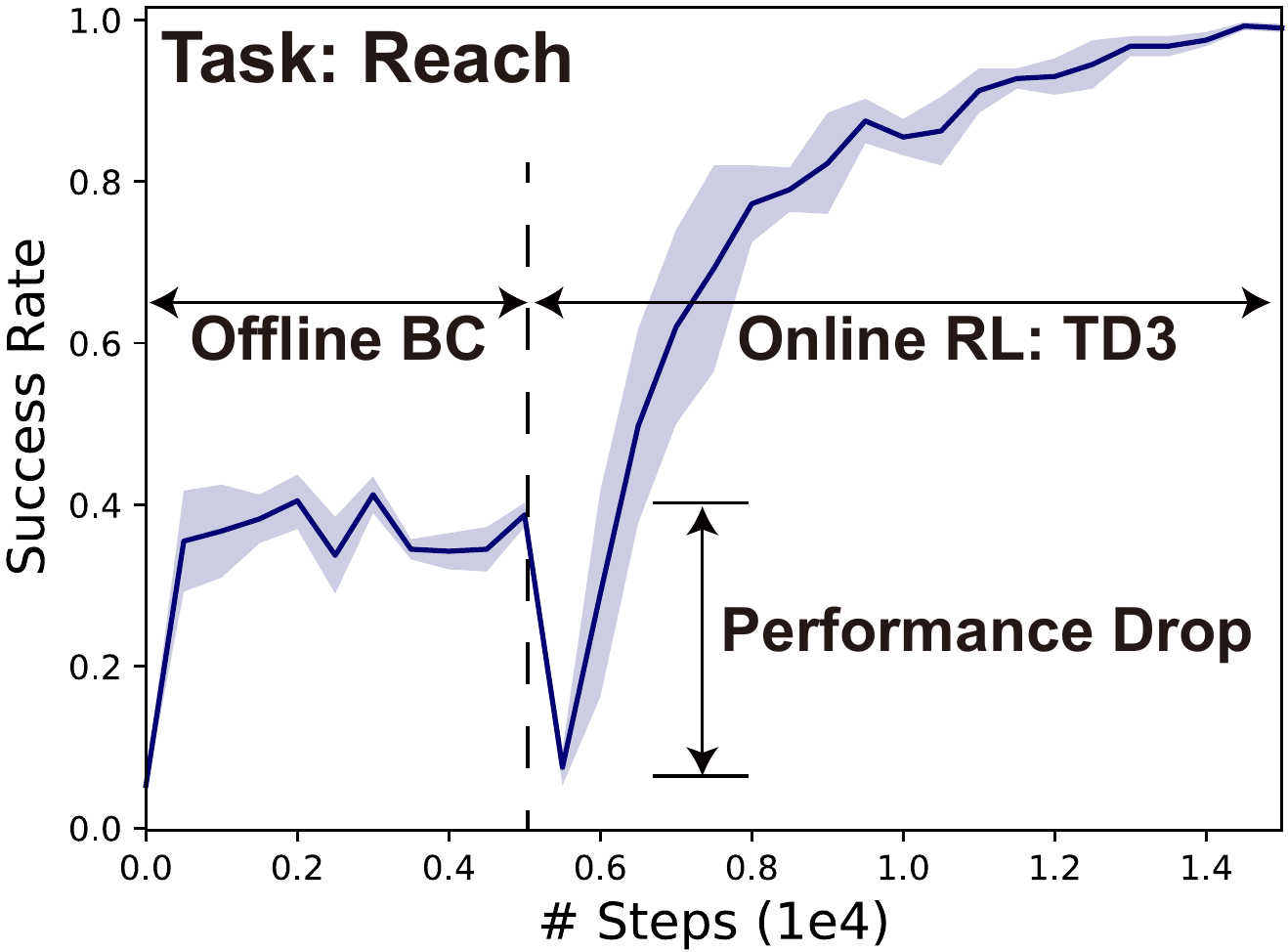}
			\label{fig:intro-3}
		}
		\caption{Impacting issues in vision-based robotic manipulation. All tasks are the OpenAI Gym Fetch suite \cite{brockman2016openai}. Results in (a) and (b) are produced using FERM \protect\cite{zhan2020framework}.
		}
	\end{figure}

Inspired by existing studies and our empirical findings, we draw several new insights toward designing highly data-efficient robotic control RL algorithms.
% Inspired by existing studies, a few insights can be draw to design highly data-efficient robotic control RL algorithms. 
First, informative state representations, including representations learned from expert demonstrations, as well as state observations that encode goal-stage information plays a very important role in accelerating policy learning (see Figure \ref{fig:intro-1} and \ref{fig:intro-2}). State representations learned from expert demonstrations and non-expert data, although both produces meaningful visual representations, has distinct impact on RL learning performance (Figure \ref{fig:intro-1}). This suggests that discriminating expert experienced states over the less helping states is very beneficial. Moreover, many complex tasks involve multiple stages (e.g. move, open/raise/close the gripper etc.), let the agent know the current task-stage and goal-related information can lead to faster learning even with sparse reward (Figure \ref{fig:intro-2}).
% \textbf{[introducing reasons here: 1) discriminate expert experienced state vs. arbitrary less helping states; 2) make the agent known current task stage, achieve faster learning even with sparse reward].}
Second, use expert demonstrations to ``warm-up'' policies and Q-functions to accelerate algorithm convergence is already adopted in a number of approaches \cite{vecerik2017leveraging,rajeswaran2017learning,nair2018overcoming,zhan2020framework}.
% fully utilize the expert demonstration to ``warm-up'' policies and Q-functions to accelerate fast algorithm convergence. 
However, as we observe in empirical studies (Figure \ref{fig:intro-3}), the transition stage between offline pre-training and online RL learning is often accompanied by severe performance drop, primarily caused by distributional shift of the policy and Q-function at the beginning of online RL training. The policy and Q-function are offline trained with respect to expert data, online exploitation and exploration may result in data not in the distribution of expert data, causing exploitation error in Q-function and distorting policy learning. This phenomenon is similar to the distributional shift issue in offline RL problems \cite{kumar2019stabilizing,levine2020offline}. Thus a better and more unified approach should be introduced to abridge the online-to-offline transition without performance loss. Moreover, as a stronger alternative to imitation learning, offline RL allows performing data-driven optimization upon a static offline dataset, which can achieve higher performance as compared with imitation learning methods, e.g. behavior cloning (BC) that na\"ively imitates the data.

In this study, we propose a new unified offline-to-online RL framework, combined with goal-aware state information and unsupervised representation learning, which achieves highly efficient robotic manipulation learning with simple sparse reward. Our proposed offline-to-online transition scheme can effectively alleviate the aforementioned performance drop. With additionally introduced goal-aware state information, our approach greatly accelerate RL training process and lead to higher data-efficiency. We evaluate the proposed framework on multiple tasks from the OpenAI Gym Fetch suite. Our method outperforms the latest state-of-the-art method FERM \cite{zhan2020framework} with much higher data-efficiency and ability to better solve harder tasks with sparse reward.

% \begin{CJK*}{UTF8}{gbsn}

% 使用强化学习来完成机械臂操控任务已经成为最近的主要方法之一，其中如何提高RL agent 的训练收敛速度一直是此方向的焦点。针对收敛速度的提升，研究者提出的方法主要集中在两点：对state representation的改进以及数据利用效率的改进
% \begin{itemize}
%     \item state representation ：在端到端的RL任务中，agent 接收到的观测量通常是非常高维度的，例如一张64x64的RGB图像通常包含12288维度数据，这远远超出使用机器人关节角度表示系统状态所需要的维度，后者通常在几十维度范围。为了让RL agent能够更好的从观测中学习到有用的信息并且使得收敛速度更快，我们会使用很多方法对原始数据进行压缩，从而得到一个更低维度的状态表征，其信息量足以让我们能够对原始数据进行还原（VAE，有待查找RL应用）或者足够让算法对场景的内容进行区分和判别（对比学习，ferm）。但是这些方法都集中于对状态空间进行更好的表征，并没有将状态表示的方法根据任务本身进行调整或者优化，例如对于机械臂分阶段抓取任务和位置到达任务来说，使用这些方法得到的状态表征是相同的，这些方法并不会根据任务的复杂度将状态压缩过程区分对待。

%     \item data efficiency ：真实世界中的机器人训练过程应当尽量减少或避免随机探索，因为真实世界中的探索代价往往很高，减少探索次数与增加数据利用效率便成了之前工作的重点之一。在 online RL 问题中，为了使算法能够提高数据利用效率，一种非常有效的方法就是对replay buffer进行优先级排序，使agent在训练过程中区分关注过去的经验，例如PER优先回放、HER、关注触觉经验的CPER。但是在复杂任务中，online 学习过程仍然存在首次到达成功时间过长的问题，为了让agent在探索的初级阶段就具有任务知识，一种非常有效的方法就是基于人类示教的 offline-RL , 例如AWAC、TD3+BC 。但是这些方法并没有深入探讨如何将offline RL 和 online RL 结合起来，很多方法（例如ferm）将 offline 过程直接连接到online过程上，这样的直接结合并不能够让agent很好的保持offline过程中学习到的知识。
% \end{itemize}

% 为了解决上面提到的两个问题并且最终实现加速机械臂操控任务的训练收敛速度，我们在manipulation任务中对以上两个方面分别提出改进方法。首先我们提出一种专门针对阶段性任务的观测量选取方法 goal-aware observation ，通过将复杂任务的reward状态和observation状态绑定，我们能够使得agent自发探索reward的内在结构，从而实现更高效的训练。其次，我们提出了一种从offline-RL转换到online-RL的过渡方案，通过设计出一个模仿-探索混合策略，并且不断调整模仿训练和随机探索的占比，我们能够更好的让agent基于offline习得的知识基础上进行online探索。

% 我们的主要贡献可以总结为：

% \begin{itemize}
%     \item 提出了一种新的机制，使得RL系统从示教学习到在线探索的过程中不会丢失示教学习得到的操控技巧
%     \item 提出了一种新的观测量设计手段，能够实现在不改动算法的情况下大幅度提升训练收敛速度
% \end{itemize}

% \end{CJK*}

\section{RELATED WORK}
\subsection{Online Reinforcement Learning for Robotic Control}
Applying deep reinforcement learning for robotic control with visual inputs have received great attention and achieved some good progress in tasks such as manipulation \cite{levine2016end}, grasping \cite{levine2018learning} and locomotion \cite{schulman2015trust,schulman2017proximal}.
However, the adoption of RL on real-world robots are still bottle-necked due to the data-inefficiency and safety issues. Online RL algorithms rely heavily on interactive exploration with the environment, which poses great challenge in relative expensive real-world data collection. To remedy this difficulty, new methods have been proposed by utilizing dense or structured reward signals \cite{wu2020learning,cabi2019scaling,camacho2020disentangled},
Sim2Real techniques \cite{pinto2017asymmetric, andrychowicz2020learning, peng2018sim} or imitation learning approaches \cite{schaal1999imitation, zhang2018deep}.
More recent studies began to combine online RL with imitation learning and unsupervised representation learning when given a small-set of human demonstration data \cite{vecerik2017leveraging,rajeswaran2017learning,nair2018overcoming,zhan2020framework}. These methods enable fully utilizing information contained in the demonstration data, which greatly accelerates the policy learning and improves data-efficiency.

\subsection{Offline Reinforcement Learning}
Offline RL (also called batch RL \cite{lange2012batch}) tackles the problem of learning policies solely from offline, static datasets, which has already achieved more performant results compared with imitation learning.
A major challenge of offline RL is \textit{distributional shift} \cite{levine2020offline}, 	which occurs when the distribution induced by the learned policy deviates largely from the data distribution. Policies can make counterfactual queries on unknown out-of-distribution (OOD) actions, causing serious overestimation of Q-values during RL training updates.
Although off-policy RL methods \cite{mnih2015human,fujimoto2018addressing,haarnoja2018soft} are designed to learn from a replay buffer, they typically fail to learn solely from fixed offline dataset and still require collecting online samples for good performance. 

Existing offline RL methods address the distributional shift issue by following three main directions.
% and achieve more performant results compared with imitation learning.
% Existing offline RL methods address distributional shift by following three major direction.
Most model-free offline RL methods constrain the learned policy to be “close” to behavior policy either by deviation clipping \cite{fujimoto2019off} or incorporate additional divergence penalties (such as KL divergence or MMD) \cite{wu2019behavior,kumar2019stabilizing}. Other model-free offline RL algorithms modifies Q-function training objective to learn a conservative, underestimated Q-function \cite{kumar2020conservative,xu2021constraints}. Model-based offline RL algorithms adopt a pessimistic MDP framework \cite{kidambi2020morel}, where the rewards of model predicted samples are penalized if they have high uncertainty\cite{yu2020mopo,kidambi2020morel,zhan2021deepthermal}.

To strive for simplicity in offline RL, Fujimoto and Gu \cite{fujimoto2021minimalist} proposes a minimalist approach, TD3+BC, which simply adds a behavior cloning term to the policy update of TD3 \cite{fujimoto2018addressing}, which balances between RL and imitation learning.

\subsection{Data Augmentation and Unsupervised Representation Learning}
Data augmentation is a widely used approach for improving data efficiency. As showed in recent studies \cite{laskin2020reinforcement,kalashnikov2018qt,kostrikov2020image}, simple data manipulation on images, like rotation, cropping and flipping, can greatly improve data-efficiency and performance of vision-based RL tasks. Data augmentation also plays a key role in contrastive learning and unsupervised representation learning in computer vision \cite{henaff2020data,he2020momentum,chen2020simple}.
Contrastive learning is a kind of self-supervised learning approach. By maximizing the agreement between similar samples and the difference between dissimilar samples, contrastive learning can obtain the informative representation of samples in an unsupervised manner.
Recent studies\cite{henaff2020data,he2020momentum,chen2020simple} have shown that pre-training on unlabeled ImageNet datasets using contrastive learning sometimes achieves better results than supervised learning on downstream classification tasks. The unsupervised representation learning methods have been applied in RL tasks in several studies \cite{laskin2020reinforcement,kostrikov2020image,zhan2020framework}, which demonstrate great improvement over data-efficiency.

\section{PRELIMINARIES}
% \textbf{TODO: Xianyuan}

\subsection{Reinforcement Learning Settings}
% 	\textbf{RL}. 
We model our robotic manipulation task as a Markov decision process (MDP) process, which is represented as a tuple $(S, A, r, T, \gamma)$, where $S$ and $A$ denote the state and action set; $r(s,a)$ is the sparse reward function, which takes the value 1 if the agent reaches the goal, -1 otherwise;
$T(s'|s,a)$ denotes the transition dynamics from current state-action pair $(s,a)$ to next state $s'$
and $\gamma\in (0,1)$ is the discount factor. To enhance the data efficiency, the states $s$ considered in our problem are the state representations encoded from the raw imagery inputs using an unsupervised representation learning model \cite{laskin2020curl}, which is trained with a small set of human expert demonstration data $\mathcal{D}$. We use reinforcement learning to solve above MDP problem, with the objective to find the optimal policy $a=\pi(s)$ to maximize the cumulative expected return  $J(\pi)=\mathbb{E}_{\pi}[\sum\nolimits_{t=0}^\infty\gamma^t r(s_t, a_t)]$. 

In the commonly used actor-critic RL paradigm, one optimizes the policy $\pi(s)$ by alternatively learning a Q-function $Q^\pi_\theta$ to minimize the Bellman evaluation errors over transitions $(s,a,r,s')$ from a replay buffer $\mathcal{B}$ as 
\begin{equation}
\begin{aligned}
\label{eq:q_evaluation}
% \scalebox{0.95}{$
J(Q) &= \underset{ \substack{ s,a,r,s' \sim \mathcal{B} \\ a' \sim \pi (s') } } {\mathbb{E}}\left[\left( y - Q^\pi_\theta (s, a) \right)^{2}\right]
% y &= r+\gamma Q^\pi_{\theta'}\left(s_{t+1}, a_{t+1}\right)
% $}
\end{aligned}
\end{equation}
where $y= r+\gamma Q^\pi_{\theta'}\left(s', \pi(s')\right)$ is the target Q-value and $Q^\pi_{\theta'}$ denotes a target Q-function, which is periodically synchronized with the current Q-function.
Then, the policy is updated to maximize the Q-value, $\mathbb{E}_{a \sim \pi_\theta}\left[Q^\pi_\theta(s, a)\right]$.

% We further define the action-value function $Q^\pi(s_t,a_t)()$

% Reinforcement learning (RL) is a kind of machine learning method aimed to complete tasks of sequential nature. Generally, these process is described as Markov decision process (MDPs) (\textit{S}, \textit{A}, \textit{R}, \textit{p}, $\gamma$), with state space \textit{S}, action space \textit{A}, scalar reward function \textit{R}, transition dynamics \textit{p}, and discount factor $\gamma$. During the training time, an agent takes actions to change its state to obtain rewards and interact with the environment until the agent find a policy $\pi$ which have the maximal expected discounted return $\mathbb{E}_{\pi}$[$\sum\nolimits_{t=0}^\infty\gamma^t\textit{r}_{t+1}$]. $\mathbb{E}_{\pi}$[$\sum\nolimits_{t=0}^\infty\gamma^t\textit{r}_{t+1}$] is the expectations of cumulative rewards and $\gamma$ is discount factor which determine the importance of future reward. We use function Q to measure this objective, which is described as $Q^{\pi}(s,a)= \mathbb{E}_{\pi}$[$\sum\nolimits_{t=0}^\infty\gamma^t\textit{r}_{t+1}|s_0=s, a_0=a]$.
	
\subsection{TD3}
    % \textbf{TD3}. 
    The backbone RL algorithm used in our proposed offline-to-online framework is based on the twin delayed deep deterministic policy gradient algorithm (TD3) \cite{fujimoto2018addressing}. TD3 is an off-policy RL algorithm that learns a deterministic policy. In TD3, two critic Q-functions $Q^\pi_{\theta_1}$ and $Q^\pi_{\theta_2}$ are learned, and the target value is evaluated by taking the minimum of the two Q-functions' estimates as follows:
    \begin{align}
    y = r + &\gamma \min_{i=1,2} Q_{\theta_i}'(s', \pi(s')+\epsilon)
    \label{eq:TD3-1}
    \\
    \epsilon &\sim \mathrm{clip}\left(\mathrm{N}(0,\sigma), -c, c\right)
    \label{eq:TD3-2}
    \end{align}
    The scheme used in Equation \ref{eq:TD3-1} is termed as clipped double Q-learning, which is shown to be highly effective in reducing overestimation error in off-policy learning. This treatment is also adopted in other off-policy RL method like SAC \cite{haarnoja2018soft}. TD3 additionally    adds a small amount of random noise $\epsilon$ to the target policy, so as to smooth the value estimates and improve robustness of the learned Q-functions.
    
    Though much simpler compared with state-of-the-art off-policy RL method SAC \cite{haarnoja2018soft}, TD3 offers comparative performance against SAC. TD3 also has a huge advantage of learning a deterministic policy as well as Q-functions less prune to value overestimation, which can be easily extended into a offline RL algorithm \cite{fujimoto2021minimalist} as will be discussed in the following content. This offers an opportunity to develop an unified framework that combines offline and online learning.
    By contrast, SAC learns an stochastic policy and solves a maximum entropy RL problem by adding an entropy term in the Q-value estimates. Although such a treatment can improve exploration in online RL, incorporating an entropy term and maximize over it can be very harmful in offline RL, as it could introduce more potential OOD errors as well as severe distributional shift.
    
    % that jointly learns an action-conditioned state value function through Q learning and a deterministic policy by maximizing expected returns.As an actor-critic method, TD3 learns an actor policy $\pi_{\theta}$ and an ensemble of critics $Q_{\phi1}$ and $Q_{\phi2}$. Specifically, the loss function of actor network is shown in Equation 1.
    % \begin{equation}
    % L(\theta) = -{ \mathbb{E}_{a\sim\pi}}[Q_{\pi} (S, A) ]
    % \end{equation}
    % where $A(\textit{S},\xi) \sim \rm {\tanh}(\mu_{\theta}(S)+\sigma_{\theta}\odot\xi)$, $\xi \sim N(0,I)$ is a sample from a normalized Gaussian noise vector. Simultaneously to actor network, TD3 also trains the critics network $Q_{\phi 1}$ and $Q_{\phi2}$ to minimize the Bellman equation in Equation 2.
    % \begin{equation}
    % L(\theta,B) = {\mathbb{E}}_{t\sim B}[(Q_{\pi} (S, A) -  (R + \gamma (1-d) min Q_{\phi i}(S^{\prime}, A^{\prime})))^2]
    % \end{equation}
    % Here, a transition $t = (S, A, S^{\prime} ,R, d)$ is sampled from the replay buffer $B$, where $(S,S^{\prime})$ are consecutive timestep observations, $A$ is the action, $R$ is the reward, and $d$ is the terminal flag.

\subsection{Offline RL via TD3+BC}
    % \textbf{TD3+BC}. 
    As an extension of TD3, Fujimoto and Gu \cite{fujimoto2021minimalist} proposes a minimalist algorithm for offline RL, called TD3+BC, which simply adds a behavior cloning regularization term to the policy update of TD3:
    \begin{equation}
    \pi = \arg\max_\pi \mathbb{E}_{(s,a)\sim \mathcal{D}}\big[\lambda Q(s, \pi(s))-(\pi(s)-a)^2\big]
    \end{equation}
    with $\lambda = \alpha/[\sum\nolimits_{(s,a)}|Q_{\pi} (s, a)|/N]$ and action ranges are set as $[-1,1]$. TD3+BC uses the BC term to force the policy do not deviate too much from the behavior policy of the offline dataset, while allowing the policy to be optimized with respect to the Q-function. The parameter $\alpha$ is used to control the strength of the regularize. A larger $\alpha$ will make the algorithm approach more RL ($\alpha=4$), while a small $\alpha$ will favor more imitation ($\alpha=1$).

    % With the aim of improving efficiency of reinforcement learning from examples, behavior cloning(BC) has been used as a regularization for policy optimization with TD3\cite{fujimoto2021minimalist}.The experiments proved that adding a behavior cloning term to the policy update of TD3 algorithm and normalizing the data is a simple but effective method to match the performance of state-of-the-art offline RL algorithms. Compared with TD3, the only change of TD3+BC is shown in Equation 3.
    % \begin{equation}
    % L(\theta) = -{ \mathbb{E}_{a\sim\pi}}[\lambda Q_{\pi}(S, A) - (A-a)^2]
    % \end{equation}
    % where $\lambda$ is a normalization term based on the average absolute value of Q which can be described as: $\lambda = \frac{\alpha}{\frac{1}{N}\sum\nolimits_{(S,a)}|Q_{\pi} (S, a)|}$, $a$ is the action stored in replay buffer.

\section{METHOD}
% Our proposed framework contains three key elements. We first describe the benefit of involving goal-aware state information in data-efficient robotic manipulation. We highlight the importance of using expert demonstrations for representation learning as well as introducing task-related goal-aware observations. Similar to other data-efficient frameworks \cite{henaff2020data,he2020momentum,chen2020simple,laskin2020curl}, we use a contrastive learning-based unsupervised representation learning model \cite{laskin2020curl} to obtain refined state representation for accelerated RL training. Finally, we develop an unified RL framework to combine offline and online learning without experiencing performance drop.
% % to resolve the performance drop issue between offline and online RL policy learning.

\subsection{Goal-aware State Information}
\subsubsection{Importance of representation learning with expert data}
Based on our empirical observations, the state representations obtained using unsupervised representation learning from expert demonstrations and non-expert data, although both provide meaningful visual representations, have huge performance differences on robotic manipulation tasks (see Figure \ref{fig:intro-1} and Section \ref{sec:ablation_GSI}). State representations learned from expert data greatly accelerates the convergence speeds compared with using non-expert data. A possible explanation could be that unsupervised representation learning via contrastive learning tend to overfit expert data, producing somewhat different state representations for expert experienced states and uncovered states, which allows the RL agents to easily discriminate different types of states during training.
Since expert experienced states correspond to high-reward, successful trajectories, this in fact enforces an expert-imitative behavior during RL training, which results in accelerated learning in robotic manipulation tasks.

\subsubsection{Goal-aware observations}
Many robotic manipulation tasks contain complex internal structures that can be seen as ensembles of a series ordered sub-tasks, with each corresponds to a stage and sub-goal.
Solving such tasks using sparse reward is quite challenging, as the reward signal it self does not contain any task structure information.
% Solving the task using a sparse reward based on final trial success is quite challenging.
% robotic manipulation tasks often comes with sparse reward and sequential structure , which could be seen as an ensemble of many sub-tasks with a sub-goal for each of their own. 
Given that vision-based robotic manipulation tasks often correspond to partially observable Markov decision processes, if involving a goal-stage aware observation that contains the information reflecting the current stage of the task, it could provide important supplementary information. This insight is also observed in several studies that exploit specially designed reward structures to enable accelerated learning \cite{cabi2019scaling,camacho2020disentangled}. An example for this is the use of hand-eyed view camera \cite{levine2018learning} for robotic grasping tasks, which has been shown to greatly improve the reinforcement learning performance.

% it is possible to involve an information observation view that contains 
% For sequential tasks, we can manually design an observation which is composed of ordinary state information and information to represent current stage of the task, or we can choose the observation that only includes state space information . It is very hard for an RL agent to learn a discrimination of task stage from raw state observation, which could be easily learned from the goal-stage-combined-observation . With goal-stage representation combined in observation as input an RL agent could converge quickly to optimal policy, which is supported in former work Reward Machine.

% one direct approch is In-Hand-View camera , which is often mounted inside the gripper of a robotic gripper to provide a first-person's view of the gripper. We have observed a phenomenon in an end-to-end fashioned reinforcement learning manipulation system that takes both global-view image and in-hand-view image as input that suggests without in-hand-view image as input, the optimal policy could be very difficult for the agent to learn. On the other hand, it is very easy for a RL agent to learn optimal policy with in-hand-view observation. This phenomenon could be easy verified in FERM.

\begin{figure}[h]
		\centering
		\includegraphics[width=0.85\columnwidth]{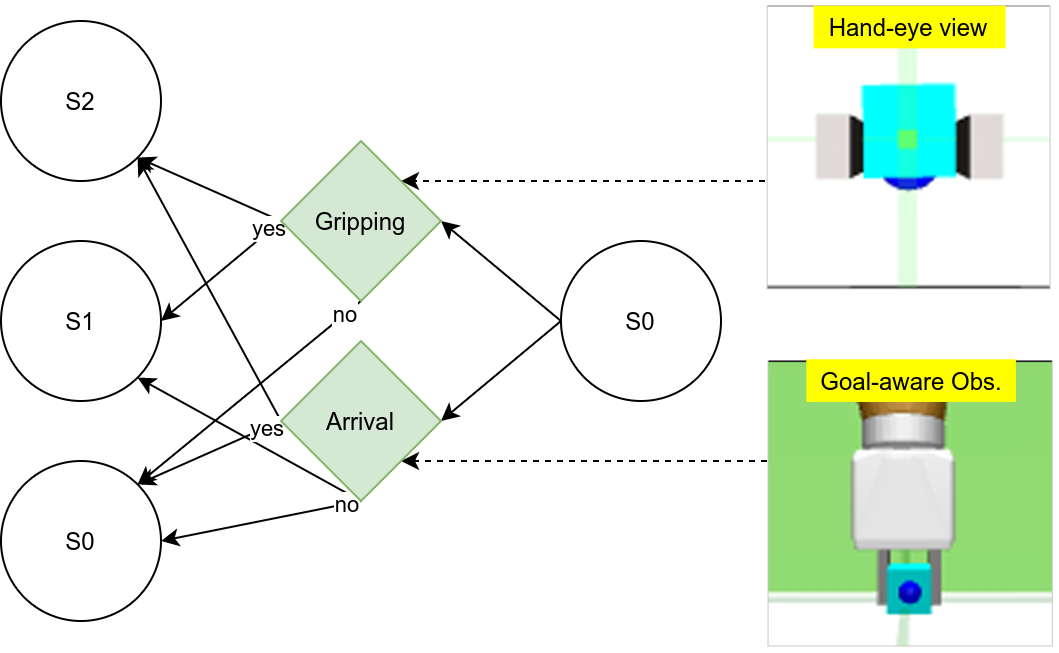}
		\caption{Stage transition logic for PickAndPlace task.
		}
		\label{fig:gsi_logic}
\end{figure}

In order to fully exploit the potential of task-stage information, we introduce additional goal-aware state information (GSI) to each task. For many robotic manipulation tasks, GSI can simply be a camera view aiming right at the goal position of the task. 
% In order to exploit the full potential of task-stage information, we introduce goal stage information (GSI)  observation. As shown in Figure \ref{fig:gsi_logic}, GSI is composed of a camera aiming right at the goal position of the current task. We simply stack this new image with global-view image and hand-eye view image as the input to an end-to-end RL system.
We illustrate this idea using a simple PickAndPlace task in Figure \ref{fig:gsi_logic}. This task can be decomposed into three stages: 1) \textit{S0}, non-gripping stage; 2) \textit{S1}, gripping but unarrival stage; and 3) \textit{S2} gripping and arrival stage. Hand-eyed view camera provides information highly related to gripping states but not providing arrival information. The global view provides overall but less specific information. In order to determine whether the robotic hand is currently in S1 or S2, one need additional goal-aware observation to reveal if the hand is arrived at target goal location. With the additional information, the task transition stage is fully specified, which can greatly reduce the learning complexity for an RL agent. In practical implementations, GSI image input can be stacked with the global view and hand-eye view images as the raw input to a end-to-end learning system, which as we show in empirical studies in \ref{sec:ablation_GSI}, can greatly improves RL training performance even with sparse reward.

% For a stage task such as PickAndPlace , we divide the task into sub-tasks like shown in Figure \ref{fig:gsi_logic} . We define the system as 3 stages non-gripping stage as S0 , gripping but unarrival stage as S1 , gripping and arrival stage as S2 . In-hand view camera provides rich information of gripping state due to an obvious reason : the content of this image is highly related to the gripping state . In order to make judgment of which state a robotic hand who has already a cube in hand is in (S1 or S2) , we need further more information that reveals if the hand is arrived at goal position . With GSI , we now have rich information that concentrates on goal arrival state representation. Ablation studies shows that RL agent could learn faster with gripping state information combined with goal state information (Figure \ref{fig:GSI_ablation}). 

\subsection{Unsupervised Contrastive Pre-training}
 To obtain the high-level state representations from raw pixel inputs, we use the unsupervised representation learning method CURL \cite{laskin2020curl}, which is the same method used in FERM \cite{zhan2020framework}. CURL uses contrastive learning that maximizes agreement between augmented data (e.g. stochastic random crop) of the same observation. Different from the original CURL model that use the stack of temporally sequential frames as an observation, we follow the treatment in FERM that use a single frame as an observation.
 We pre-train the convolutional encoder in CURL using a offline expert dataset containing 10 demonstrations. The pre-trained convolutional encoder will produce a 256-dimension state representations from the imagery inputs of all three camera angles, including the goal-aware observation. The encoded state representation are then used for RL agent training.
 More technical details about CURL can be found in previous studies \cite{laskin2020curl,zhan2020framework}.

\subsection{Offline-to-Online Reinforcement Learning}
\subsubsection{Performance drop during offline-to-online transition}
\label{sec:o2o_drop}
As mentioned in the Introduction and Figure \ref{fig:intro-3}, serious performance drop may occur when na\"ively combing an offline learning algorithm (e.g. BC or offline RL) and an online off-policy RL algorithm for training. We now elaborate the detailed mechanism of this phenomenon. Consider two offline and online learning combination schemes:
\begin{itemize}
\item Offline learning: BC; Online: off-policy RL
\item Offline learning: Offline RL; Online: off-policy RL
\end{itemize}

In offline learning stage, the former produces an imitative policy $\pi_{BC}(s)$ learned from the offline expert demonstrations $\mathcal{D}$, and the latter produces both offline learned policy $\pi(s)$ and Q-function $Q^\pi(s,a)$. The replay buffer $\mathcal{B}$ of the off-policy RL algorithm is initialized with $\mathcal{D}$ for online learning.

During online learning, the first case learns a Q-function $Q^\pi$ from the scratch using the off-policy RL. Since $Q^\pi$ is not well-learned initially, updating the policy by maximizing the Q-value $\pi=\arg\max_{(s,a)\sim\mathcal{B}}Q^\pi$ can suffer from huge performance degenerate in policy learning, even if exploring with the reasonable actions produced by the BC policy.
% relatively reasonable actions $a=\pi_{BC}(s)+\eta$ ($\eta$ is the exploration noise).

For the second case, after switching to online RL, both solely exploiting with the learned policy $\hat{a}\sim\pi$ or performing regular exploration $\hat{a}=\pi(s)+\eta$ with exploration noise $\eta$ are likely to produce state-action pairs $(s,\hat{a})$ that are not in the current replay buffer $\mathcal{B}$ in which the offline policy and Q-function are trained. Evaluation on such OOD samples will cause the similar distributional shift issue as in offline RL setting \cite{kumar2019stabilizing,levine2020offline}. As online RL lacks policy constraints or regularization that present in offline RL for combating the distributional shift, 
% Due to the lack of policy constraints or regularization in online RL, 
such error is not well-handled at the initial stage of online training, which potentially causes severe exploitation error and 
% not well-handled in online RL, causing massive exploitation and 
performance degeneration. 

\subsubsection{Unifying offline and online RL} To address the aforementioned performance drop issue and maintain the performance of the offline learned policy using expert demonstrations, we design a new unified offline-to-online RL framework based on TD3 \cite{fujimoto2018addressing} and TD3+BC \cite{fujimoto2021minimalist}.
The key idea is to maintain part of the offline RL property (i.e. constrain policy update with respect to behavioral distribution) during the offline-to-online transition phase to overcome the distributional shift issue. Formally, our offline-to-online RL framework uses the following policy update objective:
\begin{equation}
    \pi = \arg\max_\pi \mathbb{E}_{(s,a)\sim \mathcal{B}}\big[Q(s, \pi(s))- \frac{1}{\lambda} f(t) (\pi(s)-a)^2\big]
\end{equation}
Specifically, $f(t)$ is a weighting function of training step $t$ that weights the BC penalty as follows:
\begin{align}
    f(t) = 
    \begin{aligned}
        \begin{cases}
        &1, \quad\quad\quad\quad\,\,\;\mbox{ if } t\leq N_{\mathrm{off}}\\ 
        &1- \frac{t-N_\mathrm{off}}{\Delta_\mathrm{trans}}, \quad\mbox{ if } N_{\mathrm{off}} < t\leq N_{\mathrm{off}}+ \Delta_\mathrm{trans}\\
        &0, \quad\quad\quad\quad\,\,\;\mbox{ if } t> N_{\mathrm{off}}+ \Delta_\mathrm{trans}
        \end{cases}
    \end{aligned}
\end{align}
where $N_\mathrm{off}$ and $\Delta_\mathrm{trans}$ denote the training steps of offline learning and offline-to-online transition length. 
During offline learning ($t<N_{\mathrm{off}}$), $f(t)=1$ and the policy updates using the same objective as in TD3+BC; during offline-to-online transition phase,
% ($N_{\mathrm{off}} < t\leq N_{\mathrm{off}}+ \Delta_\mathrm{trans}$), 
$f(t)$ decreases linearly from 1 to 0; finally, after the transition phase ($t> N_{\mathrm{off}}+ \Delta_\mathrm{trans}$), the policy is updated normally as in online TD3.

To further reduce the potential serious OOD error caused by online exploration at the beginning of online learning, we introduce another weighting factor $g(t)$ to scale the exploration noise $\eta$ in TD3 (i.e. explore with action $a=\pi(s)+g(t)\eta$), which increase from 0 to 1 linearly during offline-to-online transition phase, and remains 1 afterwards.

Our framework perfectly merges the offline RL and online RL learning into a unified framework. And as showed in our ablation study, our framework can effectively alleviate the performance drop during the offline-to-online  transition stage. This offers several benefits: first, the expert policy recovered from offline RL stage can be maximally preserved to better facilitate online RL training; second, as we observe in empirical results, this treatment also leads to smaller variance during performance evaluation (see Section \ref{sec:ablation_o2o}).

\begin{figure}[h]
		\centering
		\includegraphics[width=0.8\columnwidth]{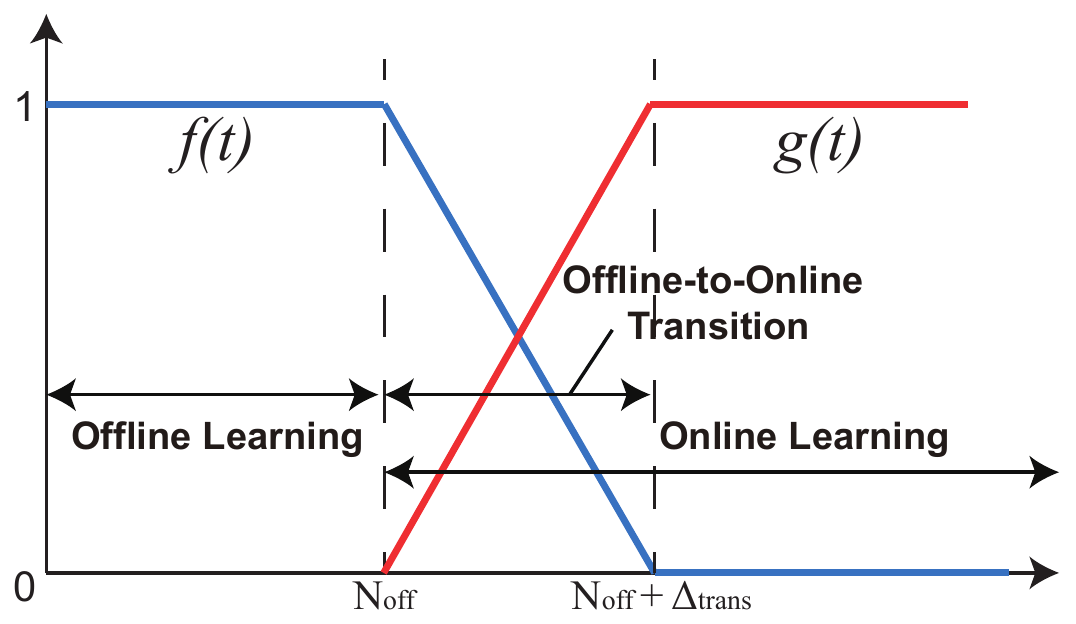}
		\caption{Illustration of the weighting function design
		}
		\label{fig:o2o_illustration}
\end{figure}

% The detailed algorithmic framework of our approach is summarized as follows:
% \begin{algorithm}
% 		\footnotesize
% 		\caption{Offline-to-online RL framework}
% 		\label{alg1}
% 		\begin{algorithmic}[1]
% 			\REQUIRE % Input
% 			Offline dataset $\mathcal{B}$
% 			\STATE XXX
% 			\STATE XXX
% 			\FOR{$XXX$}
% 			\STATE XXX
% 			\ENDFOR
% 		\end{algorithmic}
% 	\end{algorithm}

\section{EXPERIMENTS}
In this section, we present detailed experiment results and ablations of our method. We examine its training efficiency as well as the performance against state-of-the-art high data-efficiency robotic manipulation baseline FERM \cite{zhan2020framework}.

\subsection{Experiment Settings}
\begin{figure*}[t]
		\centering
		\subfloat[Reach]{
			\includegraphics[width=0.64\columnwidth]{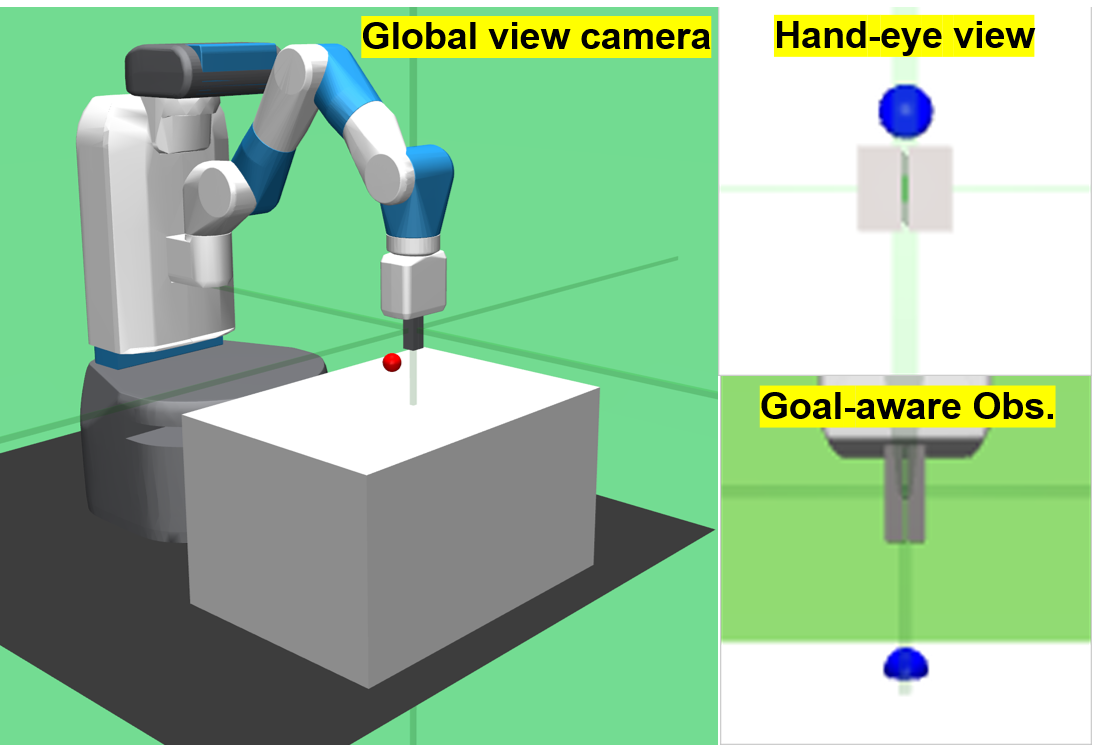}
		}\,
		\subfloat[PickAndPlace]{
			\includegraphics[width=0.64\columnwidth]{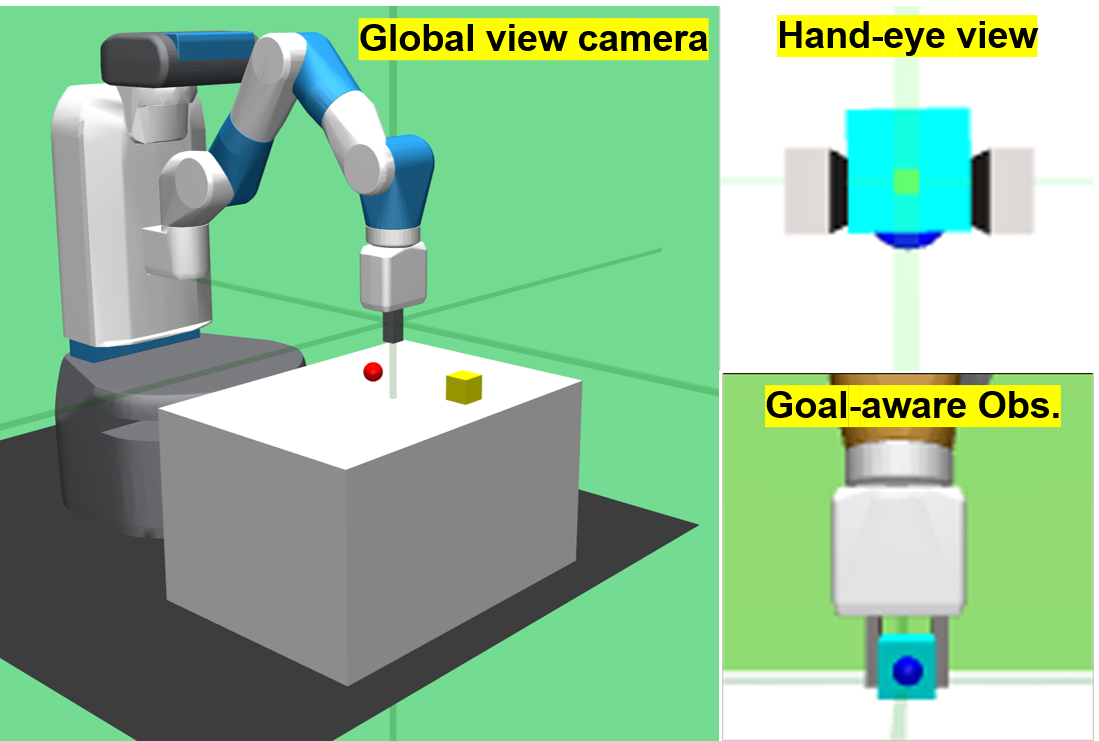}
		}\,
		\subfloat[Push]{
			\includegraphics[width=0.64\columnwidth]{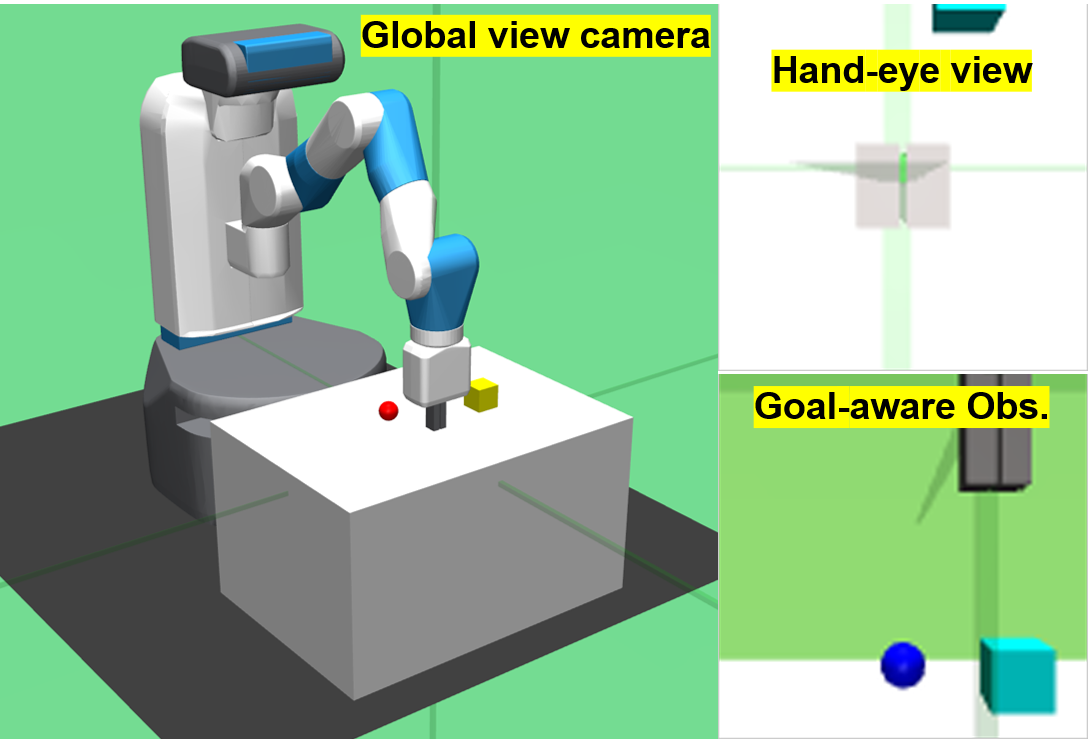}
		}
		\caption{The tasks evaluated in this work and the three observational camera inputs used in the RL agent.
		}
		\label{fig:tasks}
	\end{figure*}

\textbf{Tasks:} we evaluate our framework on three representative robotic manipulation tasks from the OpenAI Gym Fetch suite \cite{brockman2016openai}, including: 1) \textbf{Reach}: reaching an object; 2) \textbf{PickAndPlace}: picking up a block and moving it to a target location; 3) \textbf{Push}: push the block to a target location. We consider two harder tasks (PickAndPlace and Push) as compared with the tasks benchmarked in the FERM paper. PickAndPlace is a composition of Pickup and Move task in the FERM paper; and for Push, FERM fails to obtain a high performance policy.
The three tasks as well as their observational camera views for RL training are illustrated in Figure \ref{fig:tasks}. 

We train our RL framework and the FERM baseline in OpenAI Gym Fetch simulation environment. We keep the experiment and hyperparameter configurations the same as FERM, except for our TD3-based offline-to-online RL framework, as FERM uses SAC for online learning.
Both the RL algorithms of our method and FERM are trained with random crop image augmentation.

\subsection{Comparative Results}

We compare in Table \ref{tab:comparison} the results of FERM, our method, as well as a variant that without the goal-aware state information (GSI). We compare the performance of the three methods in all three tasks and investigate the training steps needed for 90\% success rate and the final performance.
% In our experiment , we compare different settings of our system to state of the art manipulation system FERM in three simulation tasks and investigate steps to 90\% success rate as well as final success rate as shown in Table \ref{tab:comparison}.

Our proposed offline-to-online RL framework with or without GSI achieves much better training efficiency compared with FERM and converge to better results.
For all three tasks, our method uses much fewer steps to reach 90\% success rate as well as the optimal policy. By applying GSI on top of our offline-to-online RL algorithm, our method achieves the best performance in most cases. The only exception is the Push task, where its convergence speed dropped slightly compared with our no-GSI variant, but still achieves better final success rate, much higher compared with the success rate of 63\% in FERM.

% our method performs the best in most tasks in simulation, except for Push task, 
% % are able to outperform most tasks in simulation, with an exception of Push task, which drops speed of convergence by a little but achieves better final success rate, the detailed reason will be discussed in Ablation Study.

% We summarize our key findings as below :

% 1. With offline-to-online transition method agents are able to find optimal policy in three tasks, include push task which converges to 60\% SR using FERM

% 2. By applying GSI on top of offline-to-online method agents are able to outperform most tasks in simulation, with an exception of Push task, which drops speed of convergence by a little but achieves better final success rate, the detailed reason will be discussed in Ablation Study.

% 3. Using state encoders trained by demonstration trajectories collected by high success rate experts greatly accelerates learning process of RL agent. 

\begin{table*}[th]
		\centering
	\caption{Performance comparison on Reach, PickAndPlace and Push Tasks. All compared methods are given 10 expert demonstrations, resulting a offline expert dataset that contains 1000 transition samples $(s,a,r,s')$.}
	\begin{tabular}{c|l|c|c|c}
		\hline
		\textbf{Tasks}	&	\textbf{Metrics}	& \textbf{Reach} & \textbf{PickAndPlace} & \textbf{Push} \\
			\hline
			&	\# steps to 90\% success rate	& 7k  & 93k & - \\
		FERM	&	\# steps to convergence & 16k & 100k & 200k \\
			&	Final success rate (\%)	&  \textbf{100} & 100 & 63 \\
			\hline
			&	\# steps to 90\% success rate	& 5k & 81k & \textbf{90k} \\
		Ours without GSI	&	\# steps to convergence	& 12k & 88k & \textbf{100k} \\
			&	Final success rate (\%)	& \textbf{100} & 96 & 88 \\
			\hline
% 			&	\# steps to 90\% SR	&  &  &  \\
% 		Ours w/o O2O	&	\# steps to convergence	&  &  &  \\
% 			&	Final SR (\%)	&  &  &  \\
% 			\hline
			&	\# steps to 90\% success rate	& \textbf{4k} & \textbf{42k} & 97k \\
			Ours	&	\# steps to convergence	& \textbf{10k} & \textbf{52k} & 110k \\
			&	Final success rate (\%)	& \textbf{100} & \textbf{100} & \textbf{91} \\
		\hline
	\end{tabular}
	\label{tab:comparison}
\end{table*}

\subsection{Ablation Study}
\subsubsection{Impact of the goal-aware state information}
\label{sec:ablation_GSI}

To study the different impact between the state representations learned from a expert demonstration (success in every trial) and non-expert ($\sim$20\% success rate) data. We evaluate our method with expert/non-expert demonstrations as well as with/without GSI. We plot the success rate evaluation results during the training process with our framework with 3 random seeds.

\begin{figure}[h]
		\centering
		\includegraphics[width=0.9\columnwidth]{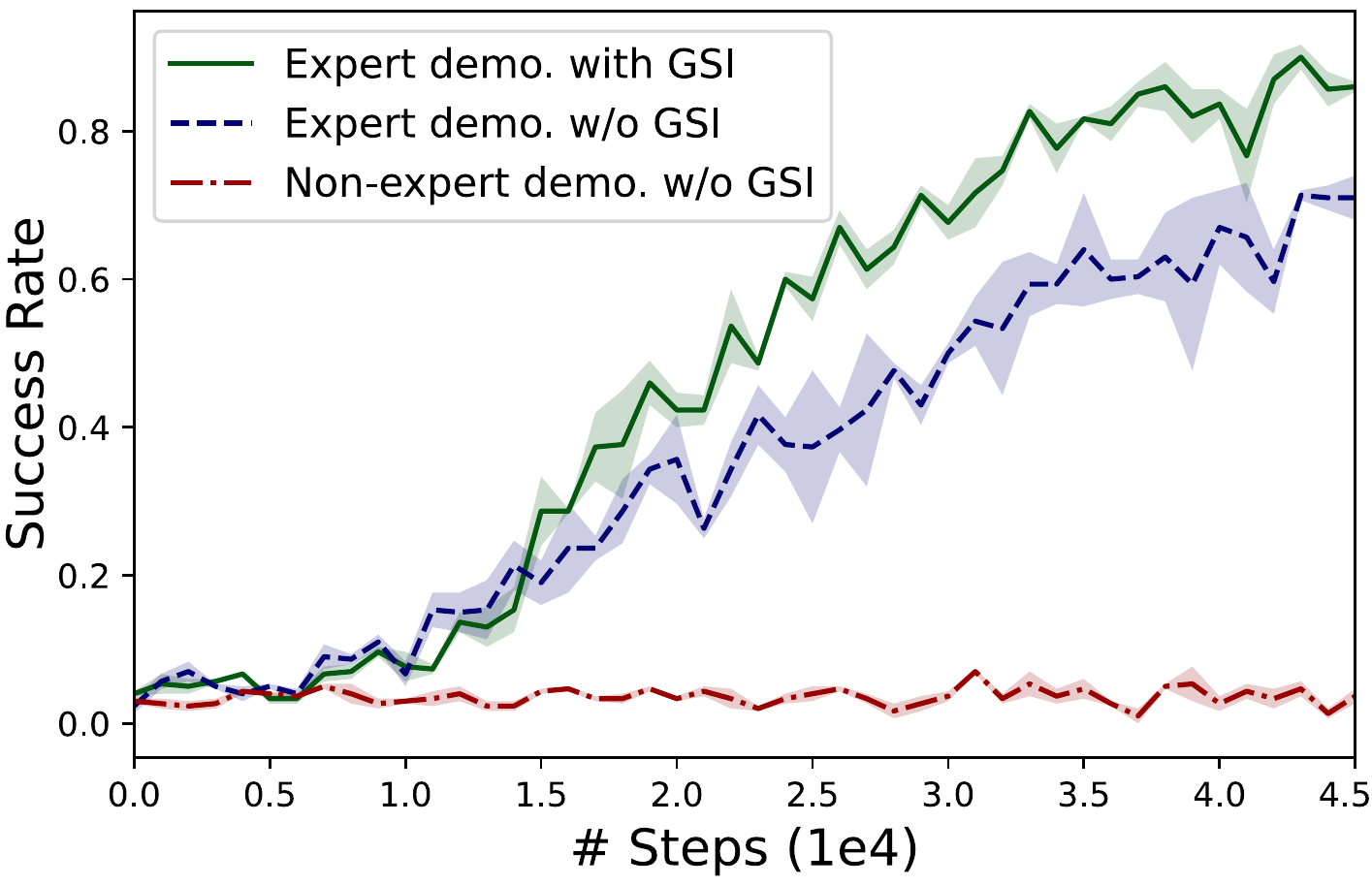}
		\caption{Ablation study on the use expert/non-expert (20\% success rate) demonstrations, as well as with/without goal-aware observations on PickAndPlace task.
		}
		\label{fig:GSI_ablation}
	\end{figure}

We find in Figure \ref{fig:GSI_ablation} that without expert demonstration and GSI, the algorithm failed to converge to any useful policy. However, with expert demonstrations, our RL agent is able to converge very quickly. Adding GSI can further accelerate convergence.

% a non-expert trained visual encoder and an  expert trained visual encoder , firstly we generate demonstration trajectories using an expert of 20\% success rate to train the encoder of our RL agent , afterwards we freeze encoder parameter and start RL training. 
% After 3 complete training episodes using different random seeds we found that the system failed to converge to any useful policy. 

% Then we generate demonstration trajectories using an expert of 100\% success rate to train the same encoder with identical parameter . Using encoder trained from success demonstrations,  our RL agent is able to converge to optimum very quickly as shown in Figure \ref{fig:GSI_ablation}.

% We are able to stabilize the learning curve of \textbf{Gym-FetchPush} task and  guarantee the final success rate by 100\% using this method, which has not been achieved by FERM. The Push-expert used in our test is trained by brute-force using non-expert demonstrations. We trained 4 agents in parallel and 1 out of 4 reached 100\% task success rate after 331k training steps, which is roughly 25\% success rate for training convergence.

Moreover, based on the results reported in Table \ref{tab:comparison} and Figure \ref{fig:GSI_ablation}, we find that GSI enhances both convergence and final performance in most tasks, especially in task with sequential execution property like PickAndPlace.
% We also evaluate our GSI method on three tasks in Table \ref{tab:comparison}. Success rate could gain largely in typical sequential task such as PickAndPlace as shown in Figure \ref{fig:GSI_ablation}. 
However, when a complex task could not be well-divided into fixed-order sub-tasks such as Push, GSI method does not guarantee boost on convergence speed, but still helpful to improve the performance of the final policy.
% performance boost on convergence speed.

\subsubsection{Offline-to-Online Learning}
\label{sec:ablation_o2o}
To systematically evaluate the performance of our offline-to-online RL framework, we compare our method with three baselines, including: 1) \textit{TD3}: using off-policy TD3 for both offline and online learning; 2) \textit{BC$\rightarrow$TD3}: offline learning using BC, online learning using TD3; 3) \textit{TD3+BC$\rightarrow$TD3}: offline learning using TD3+BC, online learning using TD3.

\begin{figure}[h]
		\centering
		\includegraphics[width=0.95\columnwidth]{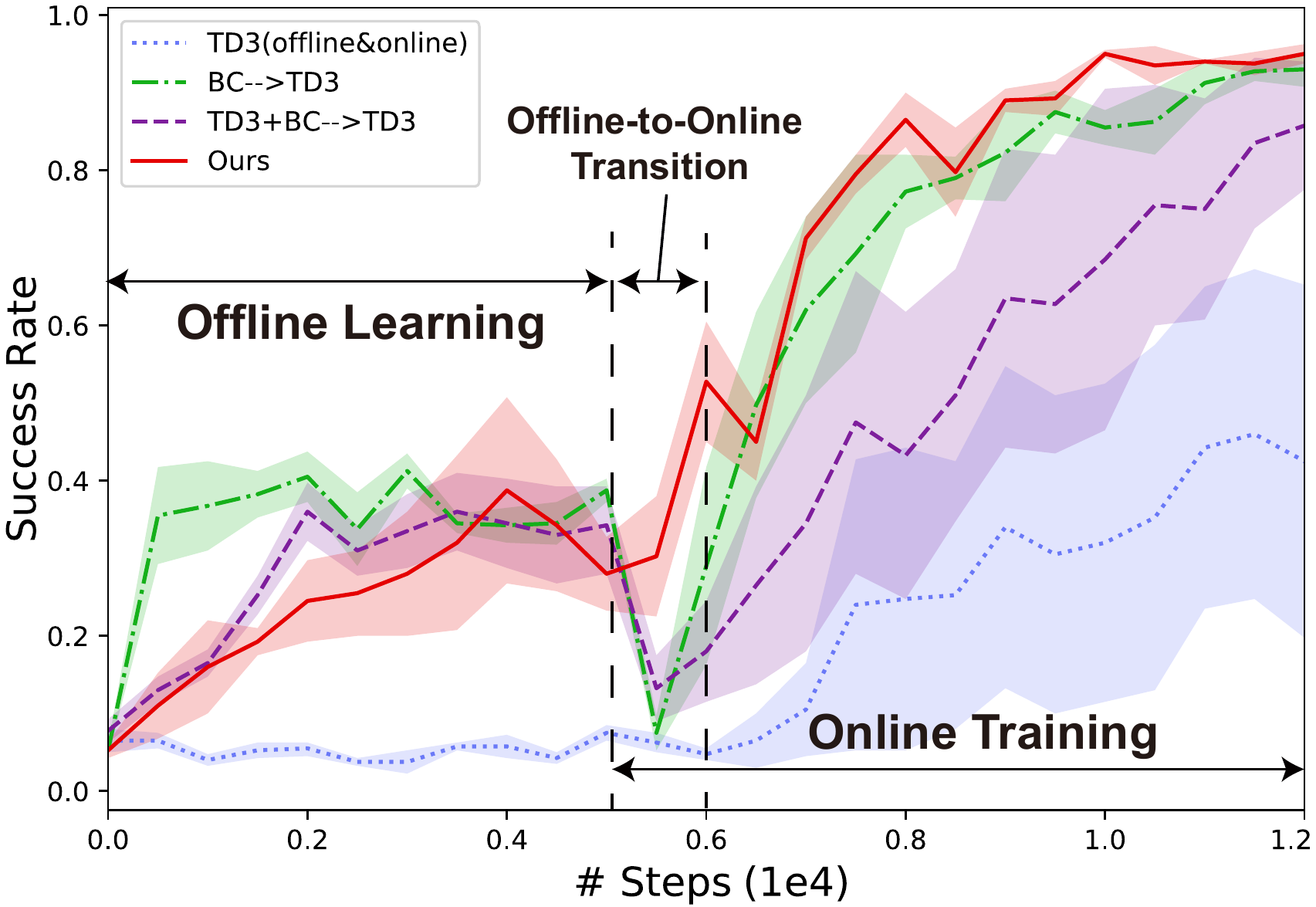}
		\caption{Evaluation of offline-to-online scheme on Reach task}
		\label{fig:off2on}
	\end{figure}

Based on the evaluation results plotted in Figure \ref{fig:off2on}, we find our approach is the only method that successfully avoids the large performance drop during offline-to-online transition phase. Our offline-to-online RL framework also achieves faster learning speed, and more importantly, has very small policy evaluation variance during the training process compared with other methods. Comparing other methods, it is observed that off-policy RL algorithm (TD3) failed during offline learning, which is as expected and also confirmed in many recent studies \cite{kumar2019stabilizing,levine2020offline}. It is also observed that na\"ively combine offline RL and online RL lead to poor performance, due to severe distributional shift at the beginning stage of online learning. Lastly, using BC for policy pre-training only and then switch to TD3 works better compared with pure TD3 and TD3+BC$\rightarrow$TD3, but still experiences large performance drop. As discussed in Section \ref{sec:o2o_drop}, this scheme is less impacted as Q-function is learned from the scratch only during online training. The performance degeneration is mainly due to optimizing with respect to the not-well learned Q-function, but suffers less severe distributional shift.

\section{CONCLUSIONS}
% \textbf{TODO: Xianyuan}
In this work, we propose a new offline-to-online RL framework for data-efficient robotic manipulation tasks. 
Our proposed method successfully resolves the performance drop issue when combining offline and online RL policy learning. Through empirical studies and mechanism analysis, we also find that the performance of expert demonstrations as well as goal-stage aware state information play an important role in enhancing RL data-efficiency and convergence performance.
% transition strategy which solves performance drop during transition of offline-RL to online -RL .Our proposed offline-to-online transition scheme can effectively alleviate the aforementioned performance drop and accelerate online learning. With additionally introduced goal-aware state information, 
We evaluate our method against the state-of-the-art baseline method as well as several variants of our approach on three robotic manipulation tasks. The experiment results show that our approach can greatly accelerate the learning process of fixed-order-decomposable manipulation tasks and achieves strong performance on all tested tasks.

\clearpage
\bibliographystyle{IEEEtran}
\bibliography{IEEEabrv,icra22}

% \clearpage
% \section*{APPENDIX}

% XXX

% \subsection{APPENDIX-1}

\end{document}